\documentclass[letterpaper]{article} 
\usepackage[preprint]{aaai2027}
\usepackage[hyphens]{url}  
\usepackage{graphicx} 
\usepackage{natbib}  
\usepackage{caption} 
\usepackage{algorithm}
\usepackage{algorithmic}

\usepackage{booktabs}
\usepackage{makecell} 

\usepackage{amsmath}
\usepackage{amssymb}

\title{When Correct Edges Cannot Be Verified: A Provenance Gap in Incomplete KGQA and a Provenance-Favoring Completion Policy}

\author{
    Yongqi Kang,
    Yu Fu,
    Yong Zhao\thanks{Corresponding author.}
}
\affiliations{
    \textsuperscript{\rm }Sichuan University\\
    2023141520239@stu.scu.edu.cn, fuyu05@stu.scu.edu.cn, yong.zhao@scupi.cn
}

\begin{document}

\maketitle

\begin{abstract}
Incomplete Knowledge Graph Question Answering (IKGQA) requires completing missing edges to continue reasoning. A growing line of work verifies completed edges against retrieved text, treating textual support as a proxy for edge quality. We ask a question that, to our knowledge, has not been systematically tested: does textual verifiability actually track correctness? Exploiting the gold deleted triples provided by the standard random-deletion protocol, we measure both. The finding is counterintuitive: among gold-correct completed edges, 76--96\% have no supporting passage even under exhaustive retrieval, robustly across deletion rates (20\%/40\%), datasets (CWQ/WebQSP), and relation types (structural, commonsense, long-tail). Most Freebase-style facts simply do not occur as head--tail co-mentions in text. Textual faithfulness therefore measures provenance, not correctness --- separated by a paradigm-level gap no in-corpus retrieval closes.

This reframes edge completion. Since most completed edges --- correct or not --- are causally redundant for the answer (95--97\% of correct answers do not depend on any unsupported edge), the central question shifts from ``is the edge correct?'' to ``admit or abstain under provenance uncertainty?'' Within this framing we present TGComplete, a provenance-favoring admission policy that retrieves evidence at a reasoning breakpoint, verifies a candidate through a lightweight loop, and abstains when support is absent. Against the generate-to-complete baseline GoG, it attains higher edge precision against gold (15--21\% vs 3--14\%), with no statistically detectable EM loss and 3.1--7.4$\times$ higher strict faithfulness of admitted edges --- at the cost of lower recall. We position TGComplete not as uniformly better, but as a principled point on a precision/provenance--recall trade-off, appropriate when auditability matters.
\end{abstract}


\section{Introduction}

Knowledge graphs (KGs) store world knowledge as structured triples and serve as an important external knowledge source for grounding LLM reasoning and combating hallucination. Recent KG-augmented question answering has progressed considerably: Think-on-Graph (ToG) treats the LLM as an agent that walks the graph hop by hop and dynamically selects relations \citep{sun2024tog}; subsequent work adds adaptive planning and self-correction \citep{chen2024pog}. Yet all these methods rest on an implicit premise --- that a complete, traversable path to the answer exists in the graph.

Real-world KGs are almost always incomplete. When a critical edge is missing, the reasoning path breaks midway, and such methods stall. To characterize this, Generate-on-Graph (GoG) introduced the Incomplete KGQA (IKGQA) setting: critical triples are randomly deleted from the KG, and a method must still answer over the depleted graph \citep{xu2024gog}. GoG's solution lets the LLM generate missing triples from parametric knowledge and admits them --- a paradigm we call \textbf{generate-to-complete}. Recognizing the hallucination risk this carries, a natural and increasingly common remedy is to \emph{verify} completed edges against retrieved text and admit only those with textual support. This raises a question that underlies the entire verify-before-admit strategy but, to our knowledge, has not been tested directly:

\textbf{Does textual verifiability actually track correctness?} If a completed edge can be supported by a retrieved passage, is it more likely correct --- and conversely, are edges without textual support actually wrong? The random-deletion IKGQA protocol offers a rare opportunity to answer this, because the deleted gold triples are known: we can label each completed edge as gold-correct or not, \emph{and} independently measure whether it has textual support, then compare the two.

The result is counterintuitive. Among \textbf{gold-correct} completed edges, 76--96\% have \textbf{no} supporting passage in the corpus --- and this holds even under exhaustive retrieval over the full corpus with relaxed windows (expanded retrieval recovers genuine support for only 4--14\% of them; most ``recovered'' co-occurrences are spurious and rejected by a strict entailment judge). The gap is robust across deletion rates (20\%/40\%), datasets (CWQ/WebQSP), and relation types (structural, commonsense, long-tail alike). The reason is structural: most Freebase-style facts --- CVT mediator nodes, database-style identifiers, facts stated across sentences or documents, facts about long-tail entities --- simply do not occur as head--tail co-mentions in any single passage. \textbf{Textual faithfulness thus measures provenance (can this edge be verified?), not correctness (is it true?); the two are separated by a paradigm-level gap that in-corpus retrieval does not close.}

This reframes what edge completion is about. We further find (via deterministic causal attribution) that most completed edges --- correct or not --- are causally redundant for the final answer: 95--97\% of correct answers do not depend on any unsupported completed edge. Together, these two findings shift the central question from ``is the completion correct?'' (often unverifiable, and usually causally irrelevant anyway) to \textbf{``admit or abstain under provenance uncertainty?''} --- a risk-management decision.

Within this framing we present \textbf{TGComplete}, a provenance-favoring admission policy: at a missing edge $(h, r, ?)$ it retrieves text evidence, verifies the candidate through a lightweight loop (text agreement, LLM entailment, KG conflict detection), and \textbf{abstains when support is absent}. We deliberately position TGComplete not as ``more correct'' but as one principled point on a precision/provenance--recall trade-off. Evaluated against GoG with gold triples as ground truth, TGComplete achieves higher \textbf{edge precision} (15--21\% vs 3--14\% strict-match against gold), with \textbf{no statistically detectable EM loss} (McNemar $p > 0.05$ on all four subsets) and \textbf{3.1--7.4$\times$ higher strict faithfulness} of admitted edges under an independent judge. The cost is \textbf{lower recall}: it rejects some correct-but-unverifiable edges (24--50\% of its rejections are gold-correct). Because rejected edges are mostly causally redundant, this recall cost does not translate into accuracy loss --- but we are explicit that it is a real trade-off, not a free lunch.

\paragraph{Contributions.} (1)~A measurement study revealing and quantifying a \emph{provenance gap} in IKGQA: 76--96\% of gold-correct completions are textually unverifiable, robustly across settings --- implying that text-faithfulness/NPR metrics measure provenance, not correctness. (2)~A causal attribution showing most completed edges are redundant for answers, reframing completion as risk management. (3)~TGComplete, a training-free, index-free provenance-favoring admission policy, characterized as a precision/provenance--recall trade-off with higher edge precision and no detectable accuracy loss. Code, prompts, and evaluation scripts are released.

\section{Related Work}

\subsection{KGQA and Path Retrieval}

Traditional KGQA falls into semantic parsing and information retrieval. Semantic parsers translate questions into executable logical queries (e.g., SPARQL): precise but highly sensitive to graph completeness. LLM-based path retrieval, represented by Think-on-Graph (ToG) \citep{sun2024tog}, treats the model as an agent that walks the graph and dynamically selects relations, avoiding fragile explicit queries. Subsequent work deepens this direction: Plan-on-Graph (PoG) introduces guidance, memory, and reflection mechanisms for adaptive-breadth planning and self-correction of erroneous paths \citep{chen2024pog}. These methods share the premise that a traversable answer path exists --- their capability is bounded by KG completeness. Our fundamental difference: when a reasoning path breaks at a missing edge, we do not merely walk within the KG but trigger text-grounded edge completion.

\subsection{Incomplete KGQA}

GoG is the first to systematically introduce the IKGQA setting, constructing incomplete graphs by randomly deleting critical triples and completing missing triples via the LLM's internal knowledge in the generation phase of a Think--Search--Generate framework \citep{xu2024gog}. We share GoG's IKGQA task setup and protocol; \textbf{the key difference is the completion mechanism}: GoG is generate-to-complete (parametric knowledge), while we are retrieve-to-complete (external text retrieval with independent verification). We additionally evaluate on the real-knowledge-update dataset IKGWQ (\S5.5) to verify our conclusions do not depend on the random-deletion protocol.

Several other lines tackle KG incompleteness with contrasting philosophies. DoM integrates structured and unstructured knowledge through a multi-agent debate: separate KG and RAG agents reason independently and a judge agent synthesizes their outputs, reporting strong accuracy on IKGWQ \citep{liu2025dom}. This \emph{debate-and-synthesize} stance differs from our \emph{verify-and-abstain} policy: DoM aims to maximize accuracy by fusing evidence, whereas we study when textual evidence can certify a completion at all and abstain when it cannot --- orthogonal goals, and a natural empirical comparison for future work (\S6.5). Earlier work used auxiliary text to fill KG gaps without LLMs: \citet{sun2023ace} parse question--candidate evidence into ``abstract conceptual evidence'' to capture fine-grained textual semantics for missing relations, anticipating our use of text to support edges but without the correctness-vs-verifiability distinction we draw. Finally, \citet{patidar2023answerability} study \emph{answerability} --- determining when a question is unanswerable due to KB incompleteness --- which is conceptually adjacent to our analysis of when completions are redundant or unverifiable (\S5.4): both ask what is recoverable given an incomplete graph, from the answerability and the provenance sides respectively.

\subsection{KG--Text Hybrid RAG}

Hybrid retrieval-augmented generation (RAG) combining structured KG and unstructured text is the closest line to ours. Think-on-Graph 2.0 (ToG-2) tightly couples KG- and text-based RAG: using the KG to link document entities for deep context retrieval while using documents as entity contexts for precise graph retrieval, alternating iteratively, training-free \citep{ma2025tog2}. GraphRAG builds a KG from documents for cross-document global summarization \citep{edge2024graphrag}. Chain-of-Knowledge dynamically grounds outputs over heterogeneous sources \citep{li2024cok}. HippoRAG performs memory-style retrieval via personalized PageRank over an offline-built entity graph \citep{gutierrez2024hipporag}.

We differ in three ways. First, \textbf{problem formulation}: ToG-2 and others perform answer-level iterative retrieval/fusion for an arbitrary query, whereas we perform targeted completion of a specific missing edge $(h, r, ?)$ at a reasoning breakpoint --- the target is a formalizable edge, not the whole answer. Second, \textbf{evaluation dimensions}: these methods primarily report accuracy, whereas we introduce and systematically report edge-quality metrics (strict faithfulness, text-grounded ratio, no-passage rate) and verify-reject behavior analysis, treating ``whether a completion is verified and how good the admitted edges are'' as a first-class objective. Third, \textbf{offline cost}: GraphRAG and HippoRAG require tens of hours of offline graph indexing (on our 6,040-entity corpus, HippoRAG indexing took ${\sim}28$ hours), whereas we require no offline preprocessing. We include HippoRAG and Self-RAG \citep{asai2024selfrag} as graph-RAG and reflective-RAG baselines respectively (\S5.1).

\subsection{Reflective RAG}

Self-RAG trains an LLM to adaptively decide whether to retrieve and to reflectively critique retrieved content and its own output \citep{asai2024selfrag}. Our verification loop shares the idea of critiquing retrieved content, but with a different objective: Self-RAG judges ``whether a retrieved passage supports the current generation,'' whereas we judge ``whether a candidate edge is sufficiently supported by evidence to enter KG reasoning,'' and upon rejection we abstain from completing the edge rather than rewriting a generation.

\subsection{LLM-as-Judge and Faithfulness Evaluation}

Using an LLM as evaluation judge (LLM-as-judge) is widely adopted for open-ended generation evaluation \citep{zheng2023judge}, but is known to exhibit systematic bias when judging its own outputs. We quantify this bias in IKGQA edge completion: the in-house backbone (Qwen2.5-7B) is systematically more lenient than an independent judge (DeepSeek-V3) when judging its own completions, while showing no such bias on GoG's completions (\S6.2). To avoid circularity, all strict faithfulness numbers use the independent judge DeepSeek-V3, and we report inter-judge agreement (Cohen's $\kappa = 0.554$).

\subsection{Recent Methods for Incompleteness and Verification}

Several recent methods address KG incompleteness or reasoning verification from angles distinct from our edge-level verify-reject. GraSP encodes a retrieved subgraph into soft prompts via a GNN, reducing sensitivity to missing edges without explicit completion \citep{wang2026grasp} --- complementary to ours: GraSP \emph{bypasses} missing edges, while we cautiously complete-or-reject when a specific edge is needed. SRP performs reference-guided iterative reflection over KG reasoning paths \citep{zhu2025srp}; its target is the whole path's relation selection, whereas we verify the textual evidence for a specific edge $(h, r, t)$ and reject rather than rewrite. StepChain GraphRAG builds a KG on-the-fly with BFS and a minimum-support threshold \citep{ni2025stepchain}, but targets general multi-hop text QA with an offline global index, whereas we target IKGQA breakpoint completion with no offline index. ORT uses ontology-guided reverse thinking \citep{liu2025ort}, relying on ontology constraints, whereas our verification focuses on single-edge textual evidence --- lighter but less structurally informed. In sum, these methods either avoid completion, verify paths, or rely on offline indexing/ontologies; our distinction is training-free, index-free retrieve-verify-reject targeted at the specific missing edge at an IKGQA breakpoint.

\section{Problem Formulation}

\paragraph{Incomplete KGQA (IKGQA).} Given a natural-language question $q$, a set of topic entities $\mathcal{E}_q$, and an \textbf{incomplete} knowledge graph $G' = (\mathcal{E}, \mathcal{R}, \mathcal{T}')$ where $\mathcal{T}'$ is a proper subset of the complete triple set $\mathcal{T}$ (i.e., $\mathcal{T}' \subsetneq \mathcal{T}$, with some critical triples needed to answer $q$ missing), the goal is to produce the answer entity set $\mathcal{A}$ for $q$ despite the incompleteness of $G'$.

\paragraph{Edge completion.} When reasoning at entity $h$ requires the value of relation $r$, but $G'$ contains no triple of the form $(h, r, ?)$, the reasoning encounters a \textbf{missing edge} at $(h, r, ?)$. Edge completion fills a tail entity $t$ for this missing edge, yielding a candidate triple $(h, r, t)$.

\paragraph{Two completion paradigms.} Generate-to-complete (GoG) has the LLM directly generate $t$ from parametric knowledge; retrieve-to-complete (ours) retrieves evidence from an external text corpus $\mathcal{C}$, determines $t$ from the evidence, and verifies its reliability.

\paragraph{Correctness vs verifiability.} We distinguish two properties of a candidate edge $(h, r, t)$. It is \textbf{correct} if it belongs to the complete triple set $\mathcal{T}$ (the deleted gold triple, recoverable in the random-deletion protocol). It is \textbf{verifiable} (text-supported) if some passage in $\mathcal{C}$ strictly entails it. A central question of this paper is the relationship between these two properties; \S5.3 shows they diverge sharply. We also distinguish three counts that must not be conflated: \textbf{missing-edge events} (reasoning breakpoints), \textbf{candidate completions} (edges proposed at those breakpoints), and \textbf{admitted edges} (candidates accepted into the chain). GoG admits every candidate it generates; TGComplete admits a subset of its candidates. The two methods need not face an identical candidate set, since candidates arise along each method's own reasoning trajectory; comparisons below are at the level of admitted/rejected edges and their gold/verifiability labels, not a shared fixed candidate set.

\section{Method}

\begin{figure*}[t]
\centering
\includegraphics[width=2\columnwidth]{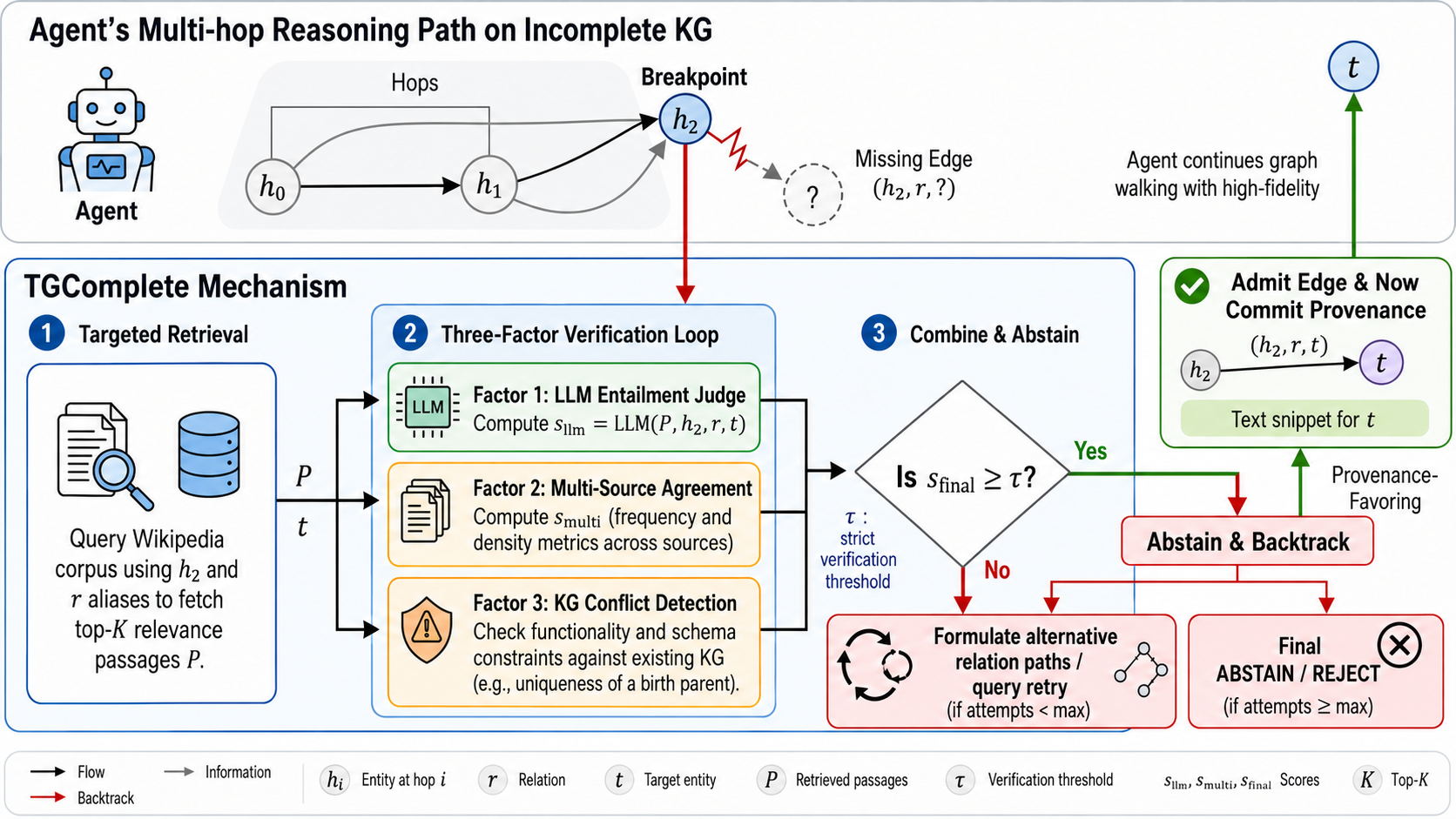}
\caption{TGComplete's retrieve-verify-abstain pipeline. At a breakpoint $(h_2, r, ?)$, TGComplete retrieves passages, computes a three-factor score ($s_{\text{llm}}$, $s_{\text{multi}}$, KG conflict), and admits the edge with provenance only if $s_{\text{final}} \geq \tau$. Otherwise it abstains and backtracks (red), or rejects permanently. The red abstention branch is the paradigm difference from generate-to-complete.}
\label{fig:framework}
\end{figure*}

TGComplete builds on a ToG-style KG-agent reasoning framework, replacing its generation phase with a retrieve-then-verify completion module that implements a provenance-favoring admission policy. The overall flow (Figure~\ref{fig:framework}): reason over the KG until encountering a missing edge $(h, r, ?)$, trigger text retrieval to obtain a candidate tail entity, verify it through the verification loop and decide to admit or abstain, then continue KG reasoning until reaching an answer.

\subsection{Targeted Text Retrieval at Breakpoints}

When reasoning breaks at $(h, r, ?)$, TGComplete constructs a query centered on $h$ and $r$ and retrieves relevant passages from the text corpus $\mathcal{C}$. The retriever combines BM25 sparse retrieval with BGE-base-en-v1.5 dense retrieval, taking the top-$k$ (default $k=5$) passages. For each entity, TGComplete maintains a text cache; when the local corpus offers no coverage, it falls back to a real-time Wikipedia lookup.

Let the retrieved passage set be $P = \{p_1, \dots, p_k\}$. TGComplete extracts a candidate tail entity $t$ from $P$, forming the candidate edge $(h, r, t)$, and records the source passages $\text{src}(t) \subseteq P$ supporting the candidate.

\subsection{Three-Factor Verification Loop}

A candidate edge $(h, r, t)$ must pass the verification loop to enter the reasoning chain. Verification comprises three factors.

\paragraph{Factor 1: Multi-source agreement.} Count the number of independent sources $n_{\text{sup}}$ supporting candidate $t$. If multiple independent passages or KG facts consistently point to the same $t$, confidence is higher:
\begin{equation}
s_{\text{multi}} = \min\left(1, \frac{n_{\text{sup}}}{n_0}\right)
\end{equation}
where $n_0$ is a saturation threshold.

\paragraph{Factor 2: LLM entailment judgment.} Invoke the backbone LLM to judge whether the candidate edge is strictly entailed by its source passages, yielding a three-way verdict --- \texttt{ENTAILED}, \texttt{UNCERTAIN}, \texttt{NOT\_ENTAILED} --- mapped to scores:
\begin{equation}
s_{\text{llm}} = \begin{cases} 1.0 & \text{\texttt{ENTAILED}} \\ \tau_{\text{unc}} & \text{\texttt{UNCERTAIN}} \\ 0.0 & \text{\texttt{NOT\_ENTAILED}} \end{cases}
\end{equation}
where the uncertain-tier score $\tau_{\text{unc}}$ defaults to 0.6.

\paragraph{Factor 3: KG conflict detection.} Check whether the candidate edge conflicts with existing facts in $G'$ (e.g., violating a relation's functional constraint). If it conflicts, reject directly.

\paragraph{Combined confidence.} Absent conflict, the combined confidence is the weighted sum of the multi-source and LLM scores:
\begin{equation}
s_{\text{final}} = w_{\text{multi}} \cdot s_{\text{multi}} + w_{\text{llm}} \cdot s_{\text{llm}}
\end{equation}
with default weights $w_{\text{multi}} = 0.35$, $w_{\text{llm}} = 0.65$.

\paragraph{Acceptance criterion.} A candidate edge is accepted into the reasoning chain if and only if there is no KG conflict and $s_{\text{final}} \geq \tau$ (default acceptance threshold $\tau = 0.5$); otherwise it is rejected (\texttt{REJECT}), and the missing edge is left uncompleted.

\subsection{Backtracking with a Fixed Threshold}

If a candidate edge is rejected, TGComplete backtracks: it re-retrieves with a more targeted query, seeking more sufficient evidence for the missing edge. A key design point: \textbf{the acceptance threshold $\tau$ stays fixed throughout backtracking, and is never lowered to accommodate a low-quality candidate.} This constraint guarantees a lower bound on completion quality --- the method would rather reject completion after multiple backtracks than admit an under-supported edge. The backtrack limit defaults to 2.

\subsection{Algorithm}

\begin{algorithm}[tb]
\caption{TGComplete's Retrieve--Verify Completion}
\label{alg:tgcomplete}
\textbf{Input}: missing edge $(h, r, ?)$, corpus $\mathcal{C}$, incomplete graph $G'$, threshold $\tau$, max backtracks $K$\\
\textbf{Output}: completed edge $(h, r, t)$ or \texttt{REJECT}
\begin{algorithmic}[1]
\STATE $k \leftarrow 0$
\REPEAT
\STATE $P \leftarrow \text{Retrieve}(h, r, \mathcal{C}, k)$ \hfill // more targeted on $k$-th backtrack
\STATE $t, \text{src}(t) \leftarrow \text{ExtractCandidate}(P)$
\IF{$\text{KGConflict}(h, r, t, G')$}
\STATE $k \leftarrow k+1$; \textbf{continue}
\ENDIF
\STATE $s_{\text{multi}} \leftarrow \min(1, n_{\text{sup}}/n_0)$
\STATE $s_{\text{llm}} \leftarrow \text{LLMEntail}(h, r, t, \text{src}(t))$
\STATE $s_{\text{final}} \leftarrow w_{\text{multi}} s_{\text{multi}} + w_{\text{llm}} s_{\text{llm}}$
\IF{$s_{\text{final}} \geq \tau$}
\STATE \textbf{return} $(h, r, t)$ with $\text{src}(t)$ \hfill // record provenance
\ENDIF
\STATE $k \leftarrow k+1$
\UNTIL{$k > K$}
\STATE \textbf{return} \texttt{REJECT} \hfill // no evidence, abstain
\end{algorithmic}
\end{algorithm}

Every accepted edge carries its source $\text{src}(t)$ (the supporting text passage IDs), logged for post-hoc auditing and our source analysis (\S5.4).

\subsection{Implementation Details}

For reproducibility and to address concerns about the concrete implementation of the verification loop, we describe each component faithfully, including its simplifications.

\paragraph{KG conflict detection.} This check targets \textbf{cardinality conflicts of functional (single-valued) relations}: we maintain a hand-curated list of functional relations (15 Freebase relations such as date of birth, place of birth, capital, spouse, plus 10 substring keywords for fuzzy matching), and reject a candidate $(h, r, t)$ when $r$ is in this list and the accumulated $G'$ triples already contain the same $(h, r)$ with a different tail $t'$. We are explicit about its boundary: the check includes \textbf{no} inverse-relation reasoning, type/range (domain-range) constraints, or learned rules; candidates with non-functional relations never trigger conflict detection. This is a deliberately lightweight implementation.

\paragraph{Text agreement (multi-source).} The multi-source score counts in how many \textbf{distinct source articles} the candidate's head and tail co-occur within a passage (the source being the Wikipedia article the passage belongs to). Deduplication is based only on passage prefix and source article name; we perform \textbf{no} semantic near-duplicate or mirror-content detection. ``Independent sources'' should therefore be read as ``co-occurrence across distinct articles,'' not independence in a strict statistical sense. Accordingly, in \S5.4 we report text-grounded ratios under a strict passage-matching protocol applied identically to both methods, avoiding the asymmetry this loose count could introduce.

\paragraph{LLM entailment judgment.} The loop calls the backbone LLM to judge whether a candidate is supported by its source passages (yes/no/uncertain, temperature $= 0$). We note the in-loop prompt does \textbf{not} hard-isolate the model's parametric knowledge, nor measure leakage. To mitigate the resulting optimistic bias, all strict-faithfulness numbers used as evidence are re-adjudicated by an \textbf{independent judge} with a stricter prompt (``mere co-occurrence without a clear relational statement is not entailment''; inter-judge agreement in \S6.2).

\paragraph{Hyperparameter setting.} The acceptance threshold $\tau = 0.5$, uncertain-tier score $\tau_{\text{unc}} = 0.6$, and weights $w_{\text{multi}}/w_{\text{llm}} = 0.35/0.65$ are \textbf{heuristic}, not tuned on any development or test set --- ruling out test-set overfitting. The sensitivity analysis in \S6.1 is a post-hoc robustness check showing the default lies on the quality--quantity Pareto frontier.

\paragraph{Candidate tail extraction (ExtractCandidate).} Given retrieved passages $P$ and missing edge $(h, r, ?)$, the backbone LLM extracts tail candidates semantically matching $(h, r)$ from each passage; candidates are ranked by number of distinct supporting source articles, and the top one is taken as $t$ with sources $\text{src}(t)$. Deduplication uses normalized entity-name matching. Error modes: extraction misses (passage has the answer but the LLM misses it --- conservative, increasing false rejections) and false positives (a near-but-wrong entity, intercepted by the subsequent entailment check). Since verification follows extraction, false positives are mostly filtered downstream.

\paragraph{Behavior on retrieval failure.} When a missing edge yields no verified candidate after backtracking, TGComplete \textbf{does not complete the edge} (\texttt{REJECT}) rather than falling back to parametric generation --- a deliberately conservative design. Allowing a low-confidence ``parametric-only'' fallback is left to future work (\S6.5).

\section{Experiments}

\subsection{Setup}

\paragraph{Datasets.} We adopt GoG's incomplete settings \citep{xu2024gog}, based on ComplexWebQuestions (CWQ) \citep{talmor2018cwq} and WebQSP \citep{yih2016webqsp}, randomly deleting critical triples with a given probability to obtain two incompleteness levels, IKG-20\% and IKG-40\%. We sample 300 questions per subset.

\paragraph{Backbone and unified settings.} The main backbone is Qwen2.5-7B-Instruct. All methods use the same backbone, the same 6,040-entity full-text Wikipedia index, and the same EM evaluation function, ensuring fairness. The corpus consists of the topic entities appearing in each CWQ/WebQSP subset plus the entities on their reasoning paths (full text, averaging ${\sim}2{,}792$ words per entity, ${\sim}16.3$M words total), covering the entities needed for reasoning under the deletion protocol; we examine its coverage ceiling and the effect of corpus expansion in the retrieval-budget sensitivity analysis of \S5.3. We use temperature $= 0$ and seed $= 42$ throughout for deterministic reproducibility. TGComplete's default configuration is in \S4.

\paragraph{Compared methods.} We compare seven methods spanning five paradigms: LLM-only (no retrieval), Vector RAG (naive RAG), Self-RAG (reflective RAG) \citep{asai2024selfrag}, HippoRAG (graph RAG) \citep{gutierrez2024hipporag}, ToG (KG path retrieval) \citep{sun2024tog}, GoG (generate-to-complete) \citep{xu2024gog}, and TGComplete (retrieve-to-complete, ours). All RAG baselines use the Qwen2.5-7B backbone and the same Wikipedia corpus.

\paragraph{Metrics.} We report: (1)~exact match (EM); (2)~\textbf{edge precision/recall against gold deleted triples} --- the fraction of admitted edges matching a gold triple, our most direct correctness measure; (3)~edge strict faithfulness --- the fraction of completed edges strictly entailed by their source passages, adjudicated by an independent judge (the judge is always a different model from the generating backbone; \S6.2); (4)~no-passage rate (NPR) --- the fraction of completed edges with no supporting passage. The contrast between (2) and (3)/(4) is itself a central object of study (\S5.3).

\subsection{Accuracy: No Detectable Difference from GoG}

Table~\ref{tab:em} reports EM for the seven methods. Two observations. First, methods that specifically handle KG incompleteness (GoG, TGComplete) vastly outperform general RAG methods --- the latter work on the simpler WebQSP (22--34\%) but systematically fail on complex multi-hop CWQ (4--8\%), since single-step retrieval cannot satisfy multi-hop compositional reasoning. Second, TGComplete and GoG are close on EM across all four subsets.

\begin{table*}[t]
\centering
\begin{tabular}{llcccc}
\toprule
Method & Paradigm & CWQ-20\% & CWQ-40\% & WebQSP-20\% & WebQSP-40\% \\
\midrule
LLM-only & No retrieval & 13.00 & 13.00 & 39.67 & 39.67 \\
Vector RAG & Naive RAG & 7.67 & 7.67 & 33.67 & 33.67 \\
Self-RAG & Reflective RAG & --- & 4.00 & --- & 22.33 \\
HippoRAG & Graph RAG & --- & 5.33 & --- & 22.00 \\
ToG & KG path & 14.67 & 10.67 & 20.33 & 20.67 \\
GoG & Generate-to-complete & 45.33 & 39.33 & 66.33 & 64.00 \\
\textbf{TGComplete} & \textbf{Retrieve-to-complete} & \textbf{45.00} & \textbf{38.33} & \textbf{68.00} & \textbf{68.33} \\
\bottomrule
\end{tabular}
\caption{Exact match (EM, \%) of seven methods on IKGQA.}
\label{tab:em}
\end{table*}

Table~\ref{tab:sig} gives the significance tests for TGComplete vs GoG. On all four subsets, the EM difference is not statistically significant (McNemar $p > 0.05$), with bootstrap 95\% CIs straddling zero. We state this carefully: \textbf{a non-significant difference means we cannot detect an EM gap at our sample size ($n=300$/subset), not that the methods are provably equivalent.} The CIs admit differences of several points in either direction (e.g., CWQ-40\% CI includes $-5.67$ pp; WebQSP-40\% shows $+4.33$ pp at $p=0.085$, a near-significant difference \emph{favoring} TGComplete). We therefore claim only \textbf{no statistically detectable EM loss} for TGComplete relative to GoG, and treat comparable accuracy as the premise for studying completion quality --- not as an equivalence claim. By contrast, the gaps of Self-RAG and HippoRAG to TGComplete are highly significant ($p < 10^{-25}$).

\begin{table*}[t]
\centering
\begin{tabular}{lcccccc}
\toprule
Subset & GoG & TGComplete & $\Delta$ (pp) & McNemar $p$ & 95\% CI & Detectable diff. \\
\midrule
CWQ-20\% & 45.33 & 45.00 & $-0.33$ & 1.000 & $[-5.0, +4.33]$ & No \\
CWQ-40\% & 39.33 & 38.33 & $-1.00$ & 0.776 & $[-5.67, +3.33]$ & No \\
WebQSP-20\% & 66.33 & 68.00 & $+1.67$ & 0.597 & $[-3.0, +6.67]$ & No \\
WebQSP-40\% & 64.00 & 68.33 & $+4.33$ & 0.085 & $[-0.01, +9.0]$ & No \\
\bottomrule
\end{tabular}
\caption{TGComplete vs GoG EM significance (McNemar test $+$ bootstrap 95\% CI).}
\label{tab:sig}
\end{table*}

\subsection{The Provenance Gap: Verifiability Does Not Track Correctness}

This section presents our central measurement finding. We use the gold deleted triples --- available because IKGQA deletes known critical triples --- to ask whether textual verifiability tracks correctness, comparing two labels for each completed edge: \emph{gold-correct} (matches a deleted gold triple or an existing KG triple) and \emph{text-supported} (has a passage that strictly entails it).

\paragraph{Edge precision against gold (Table~\ref{tab:edgeprec}).} Measured against gold triples, TGComplete admits edges of higher precision than GoG (15.2--21.4\% vs 2.6--13.5\% under strict head-relation-tail match), reflecting its abstention policy. But its recall is lower (it admits fewer edges), and 24.4--50.2\% of the edges it \emph{rejects} are in fact gold-correct --- a real cost we examine in \S5.4. This is a precision/recall trade-off, not a uniform quality win.

\begin{table}[t]
\centering
\setlength{\tabcolsep}{4pt}
\begin{tabular}{llcccc}
\toprule
Method & Subset & Adm. & \makecell{Prec.\\(strict)} & \makecell{Edge\\Recall} & \makecell{False-\\rej.} \\
\midrule
GoG & CWQ-40\% & 507 & 2.6\% & 8.8\% & --- \\
\textbf{TGComplete} & CWQ-40\% & 158 & \textbf{15.2\%} & 6.5\% & 24.4\% \\
GoG & WQSP-40\% & 370 & 13.5\% & 5.9\% & --- \\
\textbf{TGComplete} & WQSP-40\% & 257 & \textbf{21.4\%} & 4.7\% & 50.2\% \\
\bottomrule
\end{tabular}
\caption{Edge precision/recall against gold deleted triples (IKG-40\%). ``Adm.''~$=$ admitted edges; false-rejection rate is the fraction of rejected edges that are gold-correct (GoG never rejects).}
\label{tab:edgeprec}
\end{table}

\begin{figure}[t]
\centering
\includegraphics[width=0.95\columnwidth]{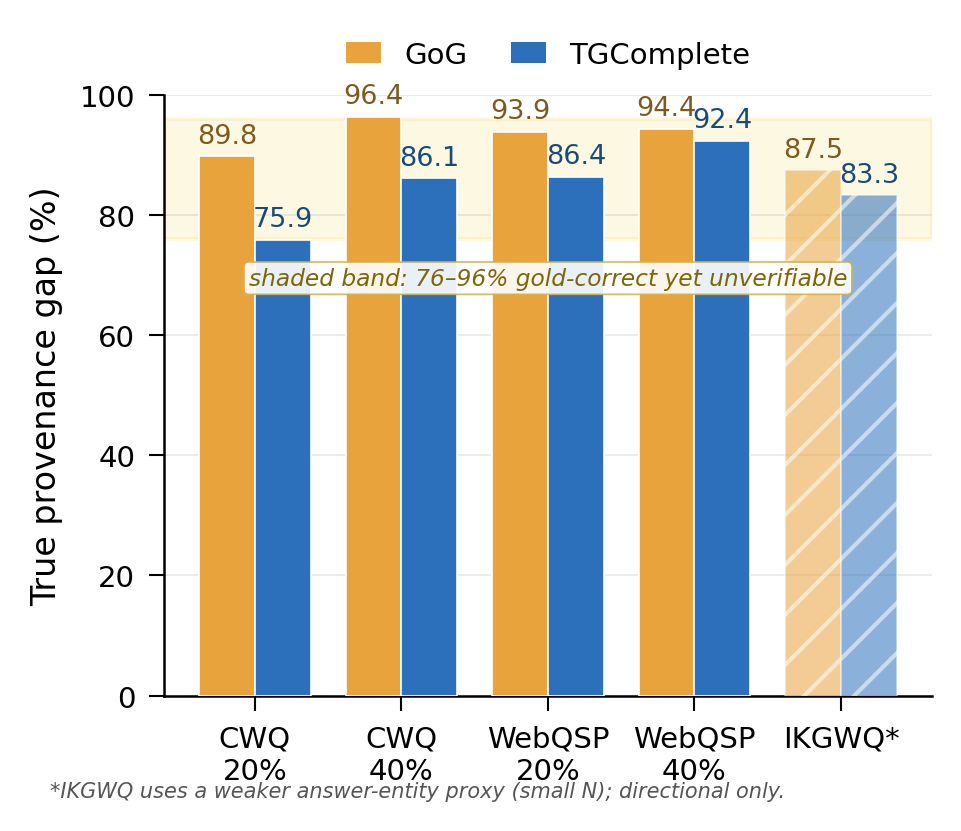}
\caption{The true provenance gap (\% of gold-correct edges that remain textually unverifiable after exhaustive retrieval) across the five settings, for GoG and TGComplete. The shaded band marks the 76--96\% range. IKGWQ uses a weaker answer-entity proxy (small $N$) and is directional only.}
\label{fig:gap}
\end{figure}

\begin{figure}[t]
\centering
\includegraphics[width=0.5\textwidth]{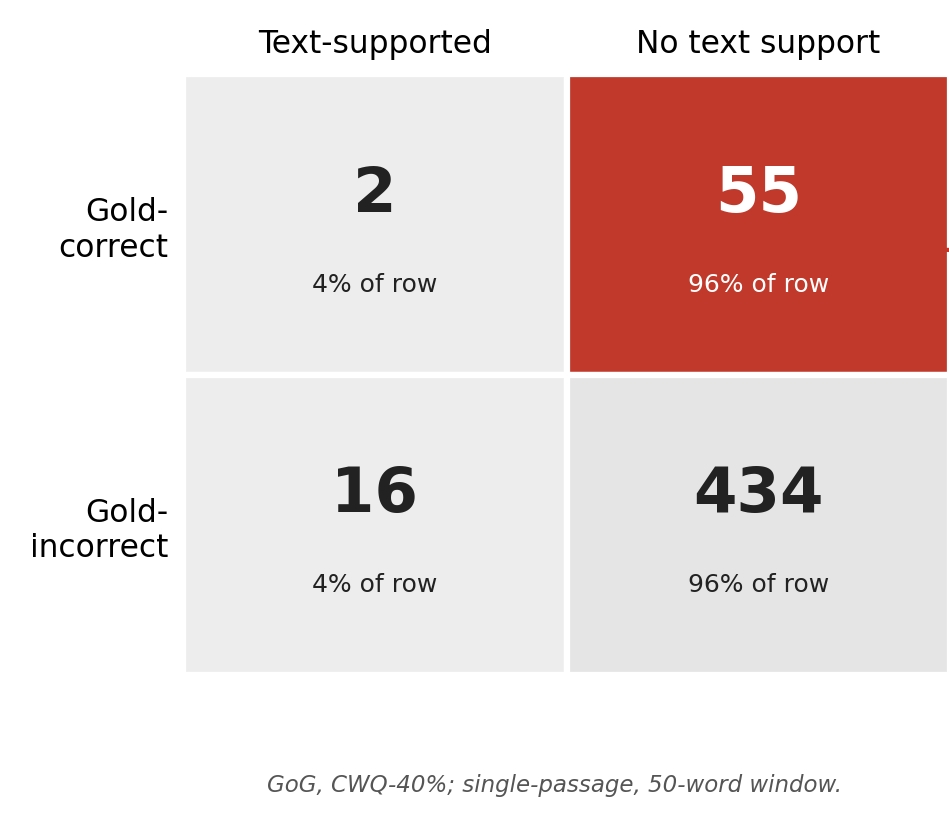}
\caption{Correctness $\times$ verifiability matrix (GoG, CWQ-40\%; single-passage, 50-word window). Among gold-correct edges, 96\% have no text support --- the highlighted ``correct-yet-unverifiable'' quadrant is the provenance gap. The same 4\%/96\% split also holds among gold-incorrect edges, confirming verifiability does not separate correct from incorrect.}
\label{fig:quadrant}
\end{figure}

\paragraph{The gap: most gold-correct edges have no textual support.} The striking finding emerges when we cross-tabulate the two labels (Figure~\ref{fig:quadrant}). Among \textbf{gold-correct} completed edges, 86--96\% have \emph{no} supporting passage under the baseline retrieval --- i.e., they are correct yet textually unverifiable. One might suspect this is a small-corpus artifact. It is not: under exhaustive retrieval (full corpus, full-text windows, with an independent entailment judge separating genuine support from spurious co-occurrence), the gap barely closes. Across gold-correct edges, the \textbf{true gap rate} --- correct edges that remain textually unverifiable even after exhaustive search --- is 76--96\%, robust across deletion rates, datasets, and relation types (Figure~\ref{fig:gap}, Table~\ref{tab:gaprobust}).

\begin{table}[t]
\centering
\begin{tabular}{llcc}
\toprule
Setting & Protocol & GoG & TGComplete \\
\midrule
CWQ-20\% & 0.2 deletion & 89.8\% & 75.9\% \\
CWQ-40\% & 0.4 deletion & 96.4\% & 86.1\% \\
WebQSP-20\% & 0.2 deletion & 93.9\% & 86.4\% \\
WebQSP-40\% & 0.4 deletion & 94.4\% & 92.4\% \\
IKGWQ & real update (proxy) & 87.5\%* & 83.3\%* \\
\bottomrule
\end{tabular}
\caption{The provenance gap is robust: true gap rate among gold-correct edges, after exhaustive retrieval. *IKGWQ uses a weaker answer-entity proxy with small $N$; reported as directionally consistent, not conclusive (\S6.4).}
\label{tab:gaprobust}
\end{table}

Expanded retrieval recovers genuine support for only 4--14\% of gold-correct no-passage edges; most ``recovered'' co-occurrences are spurious and rejected by the entailment judge (e.g., (English, spoken\_in, Cyprus) co-occurs only via ``James II of Cyprus''). The gap spans \textbf{all relation types} --- structural/CVT (94--100\%), commonsense (88--92\%), and long-tail (81--93\%) --- not just database-style relations. The reason is structural: most Freebase-style facts (CVT mediator nodes, identifier relations, cross-sentence or cross-document facts, long-tail entities) do not occur as head--tail co-mentions in any single passage.

\paragraph{Operational definition of ``exhaustive retrieval''.} For each gold-correct edge $(h, r, t)$, we search \textbf{all} 6,040 corpus articles (not only the top-$k$ from the retriever, and not only $h$'s or $t$'s own article) for any single passage in which the surface forms of both $h$ and $t$ co-occur, with the window relaxed from 50 words to the \textbf{full article text}. Any such co-occurrence is then passed to the independent entailment judge, which decides whether the passage genuinely entails $(h, r, t)$ or is a spurious co-mention. An edge counts as ``still unverifiable'' only if no article yields a judge-confirmed entailing passage. We note two boundaries this definition does not cross, both of which could in principle narrow the gap (\S6.5): (i)~it requires $h$ and $t$ to co-occur in a \emph{single} passage, so it does not perform \textbf{multi-premise aggregation} across passages or documents; (ii)~it draws evidence only from the Wikipedia-derived corpus, not from broader web text, infoboxes/tables, or KB--text alignment resources. The gap we report is therefore an upper bound on single-passage, in-corpus verifiability; whether richer evidence composition closes it is an open question we return to in \S6.5.

\paragraph{Implication.} Textual faithfulness measures \textbf{provenance} (can this edge be verified from text?), not \textbf{correctness} (is it true?). The two are separated by a gap that in-corpus retrieval does not close. Metrics like NPR and text-grounded ratio --- used here and in prior work as completion-quality signals --- should therefore be read as provenance signals. This does not make them useless: provenance is exactly what matters for auditability. But it means no text-verification method, including TGComplete, can guarantee correctness; it can only guarantee verifiability.

\paragraph{Reconciling the gap with TGComplete's precision gain.} There is an apparent tension: if verifiability does not track correctness, why does filtering for it raise edge precision against gold (Table~\ref{tab:edgeprec})? The resolution is that verifiability is a \textbf{noisy but positive} signal for correctness, not an orthogonal one --- a verified edge is more likely correct (precision rises from 3--14\% to 15--21\%), but the converse fails badly: most correct edges are \emph{not} verifiable (the 76--96\% gap). In signal terms, verifiability has reasonable precision for correctness but very low recall. This is precisely why TGComplete improves admitted-edge precision yet cannot certify correctness or achieve high recall: it keeps the high-precision verifiable slice and discards the large unverifiable-but-often-correct remainder. The precision gain is therefore real and attributable to the verification mechanism, not incidental --- but it comes bundled with the recall cost that the gap makes unavoidable.

\paragraph{Strict faithfulness of admitted edges, read as a provenance signal (Table~\ref{tab:strict}).} With this reframing, we report strict faithfulness --- the provenance quality of \emph{admitted} edges. The judge is always a different model from the generating backbone (avoiding self-judging bias; \S6.2). TGComplete's admitted edges lead GoG's by 3.1--7.4$\times$: it admits more verifiable edges by construction. This is a statement about provenance, not correctness --- consistent with the gap above.

\begin{table}[t]
\centering
\setlength{\tabcolsep}{4pt}
\begin{tabular}{lllccc}
\toprule
Backbone & \makecell{Indep.\\judge} & Subset & \makecell{GoG\\(\%)} & \makecell{TGC.\\(\%)} & Ratio \\
\midrule
Qwen2.5-7B & DeepSeek & CWQ-40\% & 2.28 & 16.77 & 7.4$\times$ \\
Qwen2.5-7B & DeepSeek & WQSP-40\% & 3.47 & 11.67 & 3.4$\times$ \\
DeepSeek-V3 & Qwen & CWQ-40\% & 4.57 & 14.13 & 3.1$\times$ \\
DeepSeek-V3 & Qwen & WQSP-40\% & 1.65 & 7.30 & 4.4$\times$ \\
\bottomrule
\end{tabular}
\caption{Strict faithfulness of admitted edges (provenance signal; independent judge, always different from the generating backbone). TGC.~$=$ TGComplete.}
\label{tab:strict}
\end{table}

The ratio is robust to retrieval budget: even under exhaustive retrieval, GoG's strict faithfulness rises only from 2.28\% to 4.14\% (CWQ) because recovered co-occurrences are mostly spurious and rejected by the judge, so TGComplete still leads (4.4$\times$ on CWQ; on WebQSP the ratio narrows to 1.7$\times$ --- an honest boundary, \S6.4).

\subsection{The Causal Role of Completed Edges: Why Completion Is Risk Management}

If most correct edges are textually unverifiable (\S5.3), what is the point of provenance-favoring completion at all? The answer is that completion quality matters less for \emph{accuracy} than one might expect --- because most completed edges, correct or not, are \textbf{causally redundant for the final answer}. This reframes completion as a risk-management decision rather than a correctness contest.

\paragraph{Most completed edges do not affect the answer.} We attribute deterministically from saved logs (no re-decoding, which would inject noise; Table~\ref{tab:causal}). For each \textbf{correct} question we reconstruct its accumulated true-edge graph $G'$ and the completion graph, and classify the answer entity's reachability: (A)~\textbf{redundant} --- reachable via $G'$ true edges alone; (B)~\textbf{dependent} --- reachable only after adding unsupported completions; (C)~\textbf{non-graph} --- supplied directly by the LLM. An unsupported edge lacks a within-50-word passage, identical for both methods.

\begin{table*}[t]
\centering
\begin{tabular}{llccccc}
\toprule
Method & Subset & Correct & (A) Redundant & (B) Dependent & (C) Non-graph & Indep.\ of unsupp.\ (A+C) \\
\midrule
GoG & CWQ-40\% & 118 & 66.1\% & 3.4\% & 30.5\% & \textbf{96.6\%} \\
GoG & WebQSP-40\% & 192 & 80.7\% & 4.7\% & 14.6\% & \textbf{95.3\%} \\
TGComplete & CWQ-40\% & 115 & 66.1\% & 4.3\% & 29.6\% & \textbf{95.7\%} \\
TGComplete & WebQSP-40\% & 205 & 79.5\% & 8.8\% & 11.7\% & \textbf{91.2\%} \\
\bottomrule
\end{tabular}
\caption{Causal attribution of correct answers (IKG-40\%; correct questions only). The last column is the share of correct answers independent of any unsupported completed edge (A+C).}
\label{tab:causal}
\end{table*}

\paragraph{95--97\% of correct answers do not depend on any unsupported completed edge.} Symmetrically, force-accepting all of TGComplete's rejected edges makes the gold answer reachable for only 2.7\% (CWQ) / 5.3\% (WebQSP) of its wrong answers --- so abstention is nearly costless for accuracy, even though (\S5.3) it discards many gold-correct edges. Those discarded correct edges are mostly redundant ones. The benefit of admitting unsupported edges is also nonzero but marginal: column (B) shows 3--5\% of GoG's correct answers do depend on an unsupported completion that happens to be correct (e.g., Australia$\rightarrow$Constitutional monarchy). Since one cannot identify in advance which unsupported edge is the lucky-correct one, admitting all of them buys a small, unpredictable accuracy contribution against a large provenance liability --- which is exactly why an abstention policy is reasonable, and why it does not cost accuracy.

A note on category (C), ``non-graph,'' which is non-trivially large (e.g., 30.5\% for GoG on CWQ-40\%): these are correct answers whose answer entity is unreachable in the reconstructed graph ($G'$ plus all completions), meaning the answer was produced directly by the agent LLM rather than by graph traversal. This is a property of the underlying ToG/GoG agent framework, which permits a final-answer step even when graph reasoning has not isolated the answer --- in practice triggered after the agent exhausts productive search/completion steps and falls back on parametric knowledge. We treat (C) as ``independent of unsupported edges'' because the answer does not flow through any completed edge; but we flag it as a distinct path --- these successes bypass the completion mechanism entirely, and a non-trivial share of both methods' accuracy reflects the backbone's parametric knowledge rather than KG reasoning. This nuance does not affect our claims (it only reinforces that completed edges, supported or not, are often not the locus of correctness), but it does mean EM partly reflects backbone knowledge, consistent with our framing of completion as risk management rather than the primary driver of accuracy.

\paragraph{Retrieval vs rejection: what drives provenance quality?} Given that abstention is the mechanism, does TGComplete's provenance advantage come from rejecting weak edges, or from retrieving better candidates? We isolate this with a \textbf{GoG+Verify} control (Table~\ref{tab:gogverify}): apply TGComplete's exact verify-reject loop (same retrieval, thresholds, judge) to GoG's \emph{generated} candidates, differing only in candidate source.

\begin{table*}[t]
\centering
\begin{tabular}{llccccc}
\toprule
Method & Subset & Admitted & Rejected & EM & strict (\%) & NPR (\%) \\
\midrule
GoG & CWQ-40\% & 507 & 0 & 39.33 & 2.28 & 89.7 \\
GoG+Verify & CWQ-40\% & 54 & 453 & 38.67 & 11.11 & 83.3 \\
\textbf{TGComplete} & CWQ-40\% & 155 & 261 & 38.33 & 16.77 & 34.8 \\
GoG & WebQSP-40\% & 370 & 0 & 64.00 & 3.47 & 84.0 \\
GoG+Verify & WebQSP-40\% & 85 & 285 & 62.00 & 3.53 & 88.2 \\
\textbf{TGComplete} & WebQSP-40\% & 257 & 232 & 68.33 & 11.67 & 46.7 \\
\bottomrule
\end{tabular}
\caption{GoG / GoG+Verify / TGComplete (separating rejection from retrieval).}
\label{tab:gogverify}
\end{table*}

\paragraph{Rejection (GoG $\rightarrow$ GoG+Verify).} Verifying GoG's generated edges rejects 77--89\% of them while EM barely moves ($-0.66$ / $-2.0$ pts), reconfirming redundancy. But admitted-edge NPR stays at 83--88\% --- \textbf{rejection alone cannot create provenance}, since LLM-generated candidates are intrinsically evidence-poor. \textbf{Retrieval (GoG+Verify $\rightarrow$ TGComplete).} With the same loop, retrieved candidates halve NPR and raise strict faithfulness --- retrieval supplies verifiable candidates that filtering generated ones cannot. Both components are needed: retrieval to source verifiable candidates, abstention to drop the rest.

\paragraph{Two negative findings, reported plainly.} (1)~KG structural corroboration over $G'$ is symmetric across methods (10--18\% vs 8--15\%; GoG even slightly higher on WebQSP), since GoG also accumulates $G'$ --- it does not distinguish the two. (2)~Answer-level human verifiability is tied: from 120 blind cards (two annotators), TGComplete and GoG are within noise (6.7\% vs 5.0\%), both low ($>93$\% not verifiable) --- consistent with the provenance gap, since end-to-end answers inherit the unverifiability of their edges. These delimit the contribution: TGComplete improves \emph{edge-level provenance and precision}, not end-to-end user-verifiability of answers.

\subsection{Robustness Across Protocols: Real Knowledge Updates}

Recent work criticizes that the random-deletion protocol may produce many inherently unanswerable questions, so accuracy drops reflect unanswerability rather than model capability. To address this, we validate on the IKGWQ dataset. IKGWQ rebuilds CWQ/WebQSP samples into multi-hop questions requiring up-to-date knowledge (e.g., 2025 facts), constructing a \textbf{real-knowledge-update} incomplete scenario rather than random deletion.

IKGWQ is highly challenging --- requiring up-to-date knowledge and multi-hop reasoning, even GoG reaches only 13.5\% EM. In this real scenario (Table~\ref{tab:ikgwq}), TGComplete shows no detectable EM difference from GoG (14.50\% vs 13.50\%, McNemar $p = 0.845$), with higher admitted-edge provenance (strict faithfulness 12.50\% vs 3.70\%; NPR 75.0\% vs 91.9\%). The ``no detectable accuracy difference, higher admitted-edge provenance'' pattern matches the deletion protocol. (Note: the provenance gap itself on IKGWQ uses a weaker answer-entity proxy, \S5.3/\S6.4; the EM and admitted-edge metrics here do not rely on that proxy.)

\begin{table}[t]
\centering
\begin{tabular}{lccc}
\toprule
Method & EM (\%) & strict (\%) & NPR (\%) \\
\midrule
GoG & 13.50 & 3.70 & 91.9 \\
\textbf{TGComplete} & \textbf{14.50} & \textbf{12.50} & \textbf{75.0} \\
\bottomrule
\end{tabular}
\caption{IKGWQ real-knowledge-update scenario (200 questions, Qwen2.5-7B).}
\label{tab:ikgwq}
\end{table}

This experiment has limitations: the absolute strict faithfulness on IKGWQ (12.5\%) is of the same order as the main deletion-protocol setting (Qwen backbone, 11.67--16.77\%) but still low, because up-to-date knowledge is hard to ground from entity-summary-based text retrieval; and the metric rests on fewer completion samples (TGComplete has 18 text-grounded edges), with weaker statistical robustness than the main experiment. Nonetheless, across two markedly different protocols, our core conclusion holds, indicating it does not depend on a specific evaluation protocol.

\subsection{Computational Cost}

TGComplete's per-question latency is about 1.6$\times$ GoG's (including retrieval and verification overhead), averaging 5.82 LLM calls per question. This extra cost buys a several-fold increase in the strict faithfulness of admitted edges and the filtering of unsupported generation. Worth emphasizing is the offline-cost contrast: the graph-RAG baseline HippoRAG required ${\sim}28.4$ hours of offline indexing on the 6,040-entity corpus (including named-entity recognition and triple extraction, ${\sim}8.3$M prompt tokens), whereas TGComplete builds no offline knowledge graph, using only the text index shared with the retrieval baselines, triggering retrieval online at inference.

\section{Analysis and Discussion}

\subsection{Sensitivity Analysis}

We examine TGComplete's robustness under 10 hyperparameter configurations (combinations of the acceptance threshold $\tau$, the uncertain-tier score $\tau_{\text{unc}}$, and the weights $w$). The default configuration ($\tau=0.5$, $\tau_{\text{unc}}=0.6$, $w=0.35/0.65$) and its neighbors all lie on the quality--quantity Pareto frontier, with stable accepted-edge counts and strict faithfulness. Raising the acceptance threshold to $\tau \geq 0.6$ makes the method more conservative (fewer accepted, higher faithfulness threshold). No configuration dominates the default on both accuracy and provenance simultaneously, indicating the default is a reasonable trade-off.

\subsection{LLM-as-Judge Bias}

We quantify LLM-as-judge bias. When the method's own backbone Qwen2.5-7B judges its own completions, its strict-faithfulness verdicts are systematically more lenient than the independent judge DeepSeek-V3's; no such bias appears on GoG's completions. Inter-judge agreement over all 1,314 verdicts is Cohen's $\kappa = 0.554$ (moderate). This shows that judging one's own outputs with one's own backbone overestimates completion quality. We therefore strictly follow the principle that \textbf{the judge is independent of the generating backbone}: the Qwen-backbone group is adjudicated by DeepSeek-V3, and the DeepSeek-backbone group by Qwen2.5-7B. This matters: the self-judging bias is substantial --- DeepSeek-V3 judges its own backbone's TGComplete completions at 33.7\% (CWQ) strict faithfulness, but an independent Qwen2.5-7B judge re-adjudicates this to 14.13\%. All numbers in Table~\ref{tab:strict} are under independent-judge adjudication, ensuring judge and generator are never the same model.

\subsection{Ablation}

We ablate the verification loop's factors on the first 150 questions (removing LLM verification, KG conflict detection, multi-source agreement, the KG channel, etc.). The EM differences of all variants from full TGComplete are not significant at $\alpha=0.05$, with removing LLM verification on WebQSP the most marginal ($p = 0.052$). This indicates the factors contribute relatively evenly to final accuracy, with no single factor dominating; consistent with our thesis, the verification loop's value is not in raising accuracy (no detectable EM difference) but in filtering unsupported completions and raising the strict faithfulness of admitted edges. The GoG+Verify three-way comparison in \S5.4 ablates at a higher level the ``retrieval'' and ``rejection'' components, showing both are indispensable.

\subsection{Limitations}

We list the main limitations. First, \textbf{the provenance gap on IKGWQ is weak evidence}: lacking deleted gold triples, we used an answer-entity proxy with small $N$, which admits some mislabeled edges; the gap is conclusively established only under the deletion protocol (CWQ/WebQSP, 20\%/40\%), and is merely directionally consistent on IKGWQ. Second, \textbf{recall cost}: TGComplete's abstention rejects 24--50\% of gold-correct candidate edges (\S5.3); although this does not reduce EM (rejected edges are mostly causally redundant, \S5.4), it means TGComplete is unsuitable when edge recall itself is the goal --- it favors precision/provenance over recall. Third, \textbf{the provenance gap bounds all text-verification methods, including ours}: since 76--96\% of correct edges are textually unverifiable, no text-grounding method can certify correctness; TGComplete certifies only verifiability. Raising absolute grounding would require corpora and evidence-linking beyond single-passage co-mention. Fourth, \textbf{backbone range}: limited to Qwen2.5-7B and DeepSeek-V3 under single-GPU resources. Fifth, \textbf{verification depth}: KG conflict detection covers only hand-curated functional relations; multi-source agreement does not model near-duplication rigorously (\S4.5). Sixth, \textbf{EM brittleness}: like prior work on CWQ/WebQSP, we use exact match, which can penalize semantically correct answers differing by alias or formatting; a semantic-aware answer metric (paired with our edge-level provenance assessment) is a worthwhile refinement. Seventh, \textbf{model scale}: our experiments use 7B-class backbones (Qwen2.5-7B) and DeepSeek-V3. Larger frontier models may store more correct facts parametrically, which could make generate-to-complete more competitive and shift the trade-off; while the provenance gap is a property of text vs KG structure (and so should persist), the relative standing of generate- vs retrieve-to-complete at frontier scale is untested here.

\subsection{Future Directions}

Our measurement raises a natural question: is the provenance gap intrinsic, or an artifact of how we look for evidence? Several directions could narrow --- or further confirm --- it. \textbf{(1) Richer evidence composition.} Our gap is measured under single-passage co-mention; multi-premise aggregation, e.g., NLI over sets of passages or cross-document evidence chains, could recover support for facts stated across sentences or documents. Extending TGComplete's loop to compose evidence over multiple passages is a concrete next step. We caution, however, that LLM entailment judges are known to over-claim support under incomplete evidence, so aggregation should be paired with conservative verification (e.g., ensembling a trained NLI model with the LLM judge) to avoid reintroducing the rationalization the strict judge currently guards against. \textbf{(2) Structured and aligned evidence.} Infobox/table views, broader web search, and KB--text alignment resources that map KB statements to byte-level textual spans could provide deterministic provenance for database-style facts that prose omits; testing whether such sources reduce the gap for structural/CVT relations would sharpen our structural explanation. \textbf{(3) Comparison with mixed-knowledge agents.} Multi-agent debate methods that integrate KG and text (e.g., on IKGWQ) target accuracy rather than provenance; a head-to-head study would clarify whether such pipelines reduce the provenance gap or merely trade it for higher recall, and where an abstention-first policy is preferable. \textbf{(4) Finer-grained gap analysis.} A per-relation-type and per-entity-frequency breakdown under exhaustive retrieval would separate CVT/identifier effects from long-tail sparsity, and a looser alias/ID-normalized matching protocol would test the robustness of the edge precision/recall figures. We see these less as patches to TGComplete than as a research program the provenance gap opens up.

\section{Conclusion}

We revisited a premise underlying verify-before-admit edge completion in IKGQA: that textual verifiability tracks correctness. Using the gold deleted triples that the random-deletion protocol provides, we showed this premise is largely false --- 76--96\% of gold-correct completed edges are textually unverifiable, robustly across deletion rates, datasets, and relation types, and not closed by exhaustive retrieval. Most KG facts simply do not exist as head--tail co-mentions in text. Text-faithfulness metrics therefore measure provenance, not correctness; the two are separated by a paradigm-level gap. We further showed that most completed edges are causally redundant for the answer, recasting completion as a risk-management decision rather than a correctness contest.

Within this reframing, TGComplete is a provenance-favoring admission policy: it retrieves evidence, verifies, and abstains under provenance uncertainty. It attains higher edge precision against gold triples and several-fold higher provenance of admitted edges, with no statistically detectable accuracy loss, at the honest cost of lower recall (it discards some correct-but-unverifiable edges). We position it not as uniformly superior but as a principled point on a precision/provenance--recall trade-off, suited to settings where auditability matters. More broadly, our measurement suggests the community should treat textual grounding as a provenance signal --- valuable for auditability but not a correctness certificate --- and design IKGQA evaluation accordingly. We release code, prompts, and evaluation scripts to support replication.

\bibliography{references}

\end{document}